\documentclass[conference]{IEEEtran}
\IEEEoverridecommandlockouts
\usepackage{cite}
\usepackage{amsmath,amssymb,amsfonts}
\usepackage{algorithmic}
\usepackage{graphicx}
\usepackage{textcomp}

\usepackage{multirow}
\usepackage{booktabs}
\usepackage{color}
\usepackage{graphicx}
\usepackage[implicit=false,
colorlinks = true,
linkcolor = blue,
urlcolor  = blue]{hyperref}
\usepackage{epsfig}
\usepackage{amsmath, amssymb}

\usepackage{capt-of}
\usepackage{multirow}
\usepackage{array}
\usepackage{caption} 
\usepackage{pslatex}
\usepackage{enumitem}

\usepackage{mathtools}
\DeclareMathOperator*{\minopt}{minimize}
\DeclarePairedDelimiter\norm{\lVert}{\rVert}%
\usepackage[dvipsnames]{xcolor}
\usepackage{tcolorbox}
\usepackage{tabularx}
\tcbset{tab2/.style={colback=SkyBlue!5!white,colframe=blue!50!black,colbacktitle=Blue!40!white,
coltitle=black,center title}}

\usepackage{colortbl}
\makeatletter
\newcolumntype{"}{@{\hskip\tabcolsep\vrule width 1pt\hskip\tabcolsep}}
\makeatother
\newcolumntype{?}{!{\vrule width 1.5pt}}

\def\BibTeX{{\rm B\kern-.05em{\sc i\kern-.025em b}\kern-.08em
    T\kern-.1667em\lower.7ex\hbox{E}\kern-.125emX}}
\begin{document}
\newcommand{\squeezeupppp}{\vspace{-8 mm}}
\newcommand{\squeezeuppp}{\vspace{-6 mm}}
\newcommand{\squeezeupp}{\vspace{-5 mm}}
\newcommand{\squeezeup}{\vspace{-3 mm}}
\newcommand{\squeezeu}{\vspace{-2 mm}}
\newcommand{\squeeze}{\vspace{-1 mm}}
\newcommand{\squeez}{\vspace{-.5 mm}}

\title{Machine Learning Subsystem for Autonomous Collision Avoidance on a small UAS with Embedded GPU
%{\footnotesize \textsuperscript{*}Note: Sub-titles are not captured in Xplore and should not be used}
\thanks{ACKNOWLEDGMENT OF SUPPORT AND DISCLAIMER: (a) This material is based upon work supported by the US Navy Contract No. N6833520C0964. (b) Any opinions, findings and conclusions or recommendation expressed in this material are those of the author(s) and do not necessarily reflect the views of the US Navy.}
}

\author{\IEEEauthorblockN{Nicholas Polosky, Tyler Gwin, Sean Furman, Parth Barhanpurkar, Jithin Jagannath}
\IEEEauthorblockA{Marconi-Rosenblatt AI/ML Innovation Lab, ANDRO Computational Solutions, LLC, Rome NY \\ \{npolosky, tgwin, sfurman, pbarhanpurkar, jjagannath\}@androcs.com
}
}

%%%%%%%%%
\iffalse
\author{\IEEEauthorblockN{1\textsuperscript{st} Given Name Surname}
\IEEEauthorblockA{\textit{dept. name of organization (of Aff.)} \\
\textit{name of organization (of Aff.)}\\
City, Country \\
email address or ORCID}
\and
\IEEEauthorblockN{2\textsuperscript{nd} Given Name Surname}
\IEEEauthorblockA{\textit{dept. name of organization (of Aff.)} \\
\textit{name of organization (of Aff.)}\\
City, Country \\
email address or ORCID}
\and
\IEEEauthorblockN{3\textsuperscript{rd} Given Name Surname}
\IEEEauthorblockA{\textit{dept. name of organization (of Aff.)} \\
\textit{name of organization (of Aff.)}\\
City, Country \\
email address or ORCID}
\and
\IEEEauthorblockN{4\textsuperscript{th} Given Name Surname}
\IEEEauthorblockA{\textit{dept. name of organization (of Aff.)} \\
\textit{name of organization (of Aff.)}\\
City, Country \\
email address or ORCID}
\and
\IEEEauthorblockN{5\textsuperscript{th} Given Name Surname}
\IEEEauthorblockA{\textit{dept. name of organization (of Aff.)} \\
\textit{name of organization (of Aff.)}\\
City, Country \\
email address or ORCID}
\and
\IEEEauthorblockN{6\textsuperscript{th} Given Name Surname}
\IEEEauthorblockA{\textit{dept. name of organization (of Aff.)} \\
\textit{name of organization (of Aff.)}\\
City, Country \\
email address or ORCID}
}
\fi
%%%%%%%%%%%%%%%%%%%%%%%%%%%%%%%%%%
\maketitle

% Marconi-Rosenblatt framework for intelligent autonomy UAV (MR-iFLY)
\begin{abstract}
Interest in unmanned aerial system (UAS) powered solutions for 6G communication networks has grown immensely with the widespread availability of machine learning based autonomy modules and embedded graphical processing units (GPUs). While these technologies have revolutionized the possibilities of UAS solutions, designing an operable, robust autonomy framework for UAS remains a multi-faceted and difficult problem. In this work, we present our novel, modular framework for UAS autonomy, entitled MR-iFLY, and discuss how it may be extended to enable 6G swarm solutions. We begin by detailing the challenges associated with machine learning based UAS autonomy on resource constrained devices. Next, we describe in depth, how MR-iFLY’s novel depth estimation and collision avoidance technology meets these challenges. Lastly, we describe the various evaluation criteria we have used to measure performance, show how our optimized machine vision components provide up to 15X speedup over baseline models and present a flight demonstration video of MR-iFLY’s vision-based collision avoidance technology. We argue that these empirical results substantiate MR-iFLY as a candidate for use in reducing communication overhead between nodes in 6G communication swarms by providing standalone collision avoidance and navigation capabilities.
\end{abstract}


\section{Introduction}
In this work, we design and demonstrate our approach to an autonomy engine deployable on a small UAS equipped with an embedded GPU-capable device. The proposed technology is a novel, general-purpose UAV autonomy framework providing robust perception, collision avoidance, and human-machine teaming for small UAS missions. The full capabilities of our framework are enabled using only basic inertial sensors and a monocular RGB camera. These capabilities are powered by both machine learning based and analytic autonomy components whose interaction provides state-of-the-art sensor enhancement and robust algorithmics. The developed framework has been successfully flight tested on commercial-off-the-shelf (COTS) quadcopters equipped with embedded GPU devices. We call the presented framework Marconi-Rosenblatt framework for intelligent autonomous UAV (MR-iFLY).

Autonomous capabilities for UAS have the potential to revolutionize a multitude of industries and applications \cite{Jagannath20UAVBook}. This potential, along with increasingly affordable costs of GPU-enabled embedded devices, has garnered significant interest in drone-powered solutions within the communications industry~\cite{ChengOJC21}.
% Accordingly, there has been extensive interest coming from both industry and academia in realizing autonomous operation of UAS. Further, the majority of such interest lies in enabling the autonomous operation of small UAS via embedded GPU devices due to the nature of intended applications and increasingly affordable costs of GPU-enabled embedded devices. 
% Industries with the greatest potential for the adoption of drone-powered solutions include insurance, transportation, and communications. 
In particular, UAS swarm and 6G technologies have great potential to impact everyday life. The higher frequencies associated with 6G technology often require line-of-sight (LOS) or minimally occluded communication links. With the current communication infrastructure consisting of base stations, these high frequencies pose major challenges for the quality of service and network coverage. To mitigate such challenges, solutions employing swarms of small UAS to provide mobile hotspots with advantageous vantage points have been proposed to provide customers with high data-rates~\cite{bertizzolo2021streaming}~\cite{Buczek21}.

While this solution provides LOS and minimally occluded links, the feasibility of such a solution remains in question due to the challenges associated with swarm operation. Importantly, it is infeasible for human operators to control swarms due to their number, thereby necessitating autonomous UAS capabilities to enable 6G swarm technologies. In addition to this obvious problem, autonomous swarms will need to possess capabilities to evade aerial obstacles and other swarm members; the latter of which has garnered solutions involving communicating and relaying swarm members’ locations across the network. Unfortunately, such network traffic reduces the bandwidth available for the customer below. 

\textbf{Contribution.} In response to the described problem, we propose a unique autonomous UAS framework, MR-iFLY, designed to operate on an embedded GPU device with a low SWaP sensor configuration. We augment state-of-the-art machine vision models via model reduction and hardware-specific optimization techniques to allow for efficient inference on embedded GPU devices. MR-iFLY subsequently employs these optimized models in conjunction with traditional analytic autonomy components to provide robust collision avoidance and navigation. Furthermore, MR-iFLY has been successfully deployed on a small UAS equipped with an embedded GPU and requires a minimal sensor configuration.

\textbf{Impact.} Robust, vision-based collision avoidance and autonomy for UAS can minimize, if not eliminate, the need for communication overhead in 6G swarm networks. Enhanced with onboard machine perception, UAS repeaters may perceive each other and environmental obstacles, and subsequently plan actions to avoid collisions while freeing bandwidth for users. Further, efficient implementations provide lower decision-making latencies preventing the vulnerability of broken links in cloud-based solutions. We believe that the work described herein is a step towards confirming the feasibility of an autonomous UAS solution which, in the future, could be extended to aid a 6G swarm solution. 

% The rest of this paper is organized as follows. We begin by discussing the design challenges we encountered and the design philosophy that we devised to meet these challenges. We go on to discuss the implementation of our framework followed by our experimental analysis of the developed framework and end the paper with a discussion of related work, future work, and concluding remarks.

\section{Challenges and Design Philosophy}
% There are numerous design choices to be made when developing a UAS autonomy framework and making such choices in and of itself is a challenging task. 
In this section, we provide a brief overview of the challenges that engineers meet when designing an autonomy framework for embedded deployment. For each challenge, we discuss how our design philosophy meets these challenges.  

\subsection{Sensor Selection}
% The primary challenge associated with UAS autonomy is that of selecting the sensors with which to equip the UAS. 
The selection of UAS sensors has implications that affect almost every other aspect of the design of the framework; the sensor selection will determine the degree to which the UAS can perceive the operational environment and the amount of computational resources required to process sensor data. In addition to sensor selection, it is important to account for particular algorithmic components that will enhance or supplement sensor readings.

We believe that the majority of the necessary environmental information can be captured using a sensor suite consisting of a monocular RGB camera, inertial measurement unit (IMU) sensors, optical flow sensors, and GPS. Each of the aforementioned sensors is monetarily inexpensive and offers high sample rates. For these reasons, we believe that the described sensor suite renders MR-iFLY an affordable and effective solution for applications such as 6G swarm technologies. We note that, should other sensors be available aboard deployment UASs, they may be used to enhance the perceptual components of MR-iFLY, but that our goal in sensor selection was to ensure universality via the framework's minimal sensor requirement. To supplement our sensor suite, we employ machine learning-based computer vision techniques to extract information from the RGB observations. The field of machine learning based computer vision is burgeoning with techniques to enhance monocular camera-based systems, allowing for the replacement of traditional sensors (such as stereo, LiDAR, etc.) with software algorithms (depth estimation). \textit{In MR-iFLY, we posit that the reduction in sensing fidelity is worth the reduction in sensing overhead, power consumption, and data processing.}

\subsection{Computational Resources}
On a small UAS with an embedded computing system, the computational complexity of the algorithmic components and data processing techniques is of paramount importance. Many applications require near real-time response times and thus each component of the data processing pipeline should be carefully engineered to operate as efficiently as possible. This often requires re-implementing or adjusting out-of-the-box or open-source software solutions to fit the computational needs of the application.

Out-of-the-box neural network models often require prohibitively long inference times to yield their outputs. Accordingly, using such models in an embedded autonomy framework requires manipulating the models to reduce latency. In MR-iFLY, we use two separate techniques to this end: model reduction via knowledge distillation and hardware-specific optimization using TensorRT. Knowledge distillation reduces model size by teaching a smaller network to mimic a larger pre-trained network. TensorRT operates by running multiple tests on the embedded hardware to determine how to optimize each network operation. In some cases, these techniques are not enough and engineers should consider training custom networks. If this latter option does not yield desirable results traditional computer vision methods may be investigated.

\subsection{Robustness}
While algorithmic efficiency is critical, algorithmic robustness is tantamount. UASs often encounter varied operational scenes and environments yielding very different sensor readings and interference conditions. Creating an overall robust solution requires ensuring each of the subcomponents of the framework is robust and for machine learning components, robustness is tightly coupled with the composition of the training data set. We believe that utilizing models which have been explicitly trained to be robust to varying scene and environmental characteristics is of key importance. Additionally, using analytic autonomy components in place of end-to-end learning components can mitigate issues associated with the lack of a robust training data set. Further, such analytic algorithms can often be explicitly redesigned by the engineer to inherently increase robustness and cautiousness.

Overall, employing both machine learning based and traditional analytic solutions for autonomy allow for reaping the benefits of both approaches. Machine learning components offer significant increases in representational power and generalization capability which undoubtedly improves the overall autonomy solution; however, it is beneficial to exploit analytic components, when possible, to reduce the complexity of the task we assign the machine learning components to learn. 
% {\color{red}A similar paradigm is seen in the modularization of the various machine learning components e.g., the separation of visual representation and reinforcement learning policy models allows for each component to learn its respective, simpler task, as opposed to requiring a single model to learn one exceedingly complex task.}

\section{Design of Marconi-Rosenblatt Framework for Intelligent Autonomous UAV (MR-iFLY)}

\begin{figure}[!htb]
    \centering
    \includegraphics[scale=0.25]{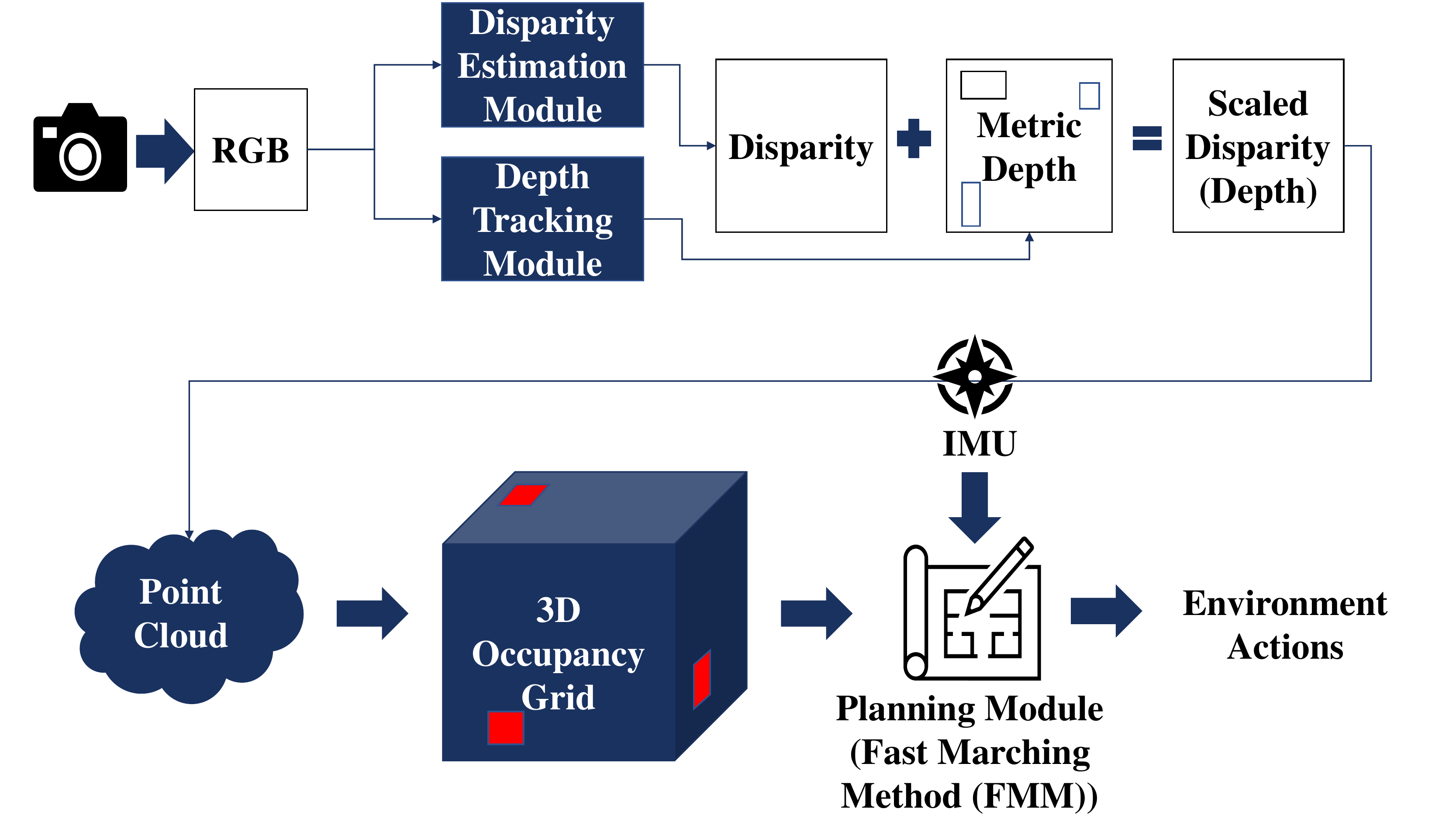}
    \caption{MR-iFLY System Diagram}
    \label{fig:sysdiag}
    %\squeezeupp
    \vspace{-4 mm}
\end{figure}

We now describe in detail the subcomponents of MR-iFLY and discuss their relationships as outlined in Figure~\ref{fig:sysdiag}.

The input into our architecture includes an RGB image collected from an FPV camera onboard the drone, and IMU readings used within the planning module. The collected RGB image is passed into two submodules: the disparity estimation module and the depth tracking module, producing, a dense estimate of image disparity and a sparse estimate of metric depth, respectively. Within the disparity module, machine learning based convolutional neural network (CNN) and visual transformer models predict disparity values for each pixel in the RGB input, where the disparity is defined as the pixel distance between the same observed scene location in the right and left images of a stereo pair. The disparity module thereby obtains an estimate of values often gathered using a stereo camera system, further providing an advantage over other autonomy solutions by reducing the sensor requirements of our architecture. Within the depth tracking module, classical computer vision methods are used to estimate a sparse metric depth map which is subsequently used to scale the dense disparity map producing a dense metric depth map.

This metric depth map is transformed into a point cloud using the camera's intrinsic characteristics and is then discretized and binned to produce the 3D occupancy grid containing binary obstacle labels. Finally, the 3D occupancy grid is passed into the planning module which outputs a sequence of UAS actions to perform in the environment. In the present architecture, we have implemented the analytic path planning algorithm based on the Fast-Marching Method (FMM) which solves for shortest time paths given a cost function. A baseline cost function considers obstacles as locations with infinite cost but more complex functions accounting for other environmental factors may be used at no further computational cost.

\subsection{Depth Estimation}

Estimating the depth of objects and obstacles in the environment surrounding the UAS is a crucial step in any motion or path planning algorithm. In traditional autonomy frameworks, this step often leverages dedicated depth sensors, such as 3D LiDAR, or utilizes a stereo camera system, from which, estimates of the metric depth can be obtained. In our framework, we employ a combination of classical computer vision and novel deep learning techniques, as described below, to estimate metric depth from monocular RGB images.

\subsubsection{Machine learning based depth estimation}

The first step MR-iFLY takes toward ascertaining metric depth - the raw distance values from a scene point to the camera lens - is to use a disparity estimation neural network. Disparity estimation networks have recently grown in popularity as the scale-invariant properties of disparity allow for networks to be trained on larger data sets whilst constraining the range of output values the network is required to learn. Accordingly, disparity estimation networks are often more robust to changes in scene and zero-shot transfer scenarios, as evidenced in \cite{Ranftl2020}. 

Despite their desirable zero-shot and robust estimation properties, network architectures such as those proposed in \cite{Ranftl2021}\cite{monodepth17} are often prohibitively large for embedded device deployment, requiring substantive inference times, even on a GPU-equipped embedded device. To reduce inference times, there exist two primary avenues outside of increasing compute resources: model size reduction and platform-specific optimizations. To reduce the size of our depth estimation model, we leverage the knowledge distillation techniques proposed in \cite{aleotti2020real}. Knowledge distillation is the process of training a smaller, student network to learn the function represented by a larger, teacher network via a semi-supervised learning scheme. To implement this scheme, we curated a data set of real-world images representing the expected distribution of scenery that our UAS solution will encounter, and supplemented this data set with images selected from the OpenImages~\cite{OpenImages} and COCO~\cite{COCO} data sets. The final data set consisted of 42,159 self-collected images, 371,159 images from OpenImages, and 75,880 images from COCO. In our knowledge distillation pipeline, we used the MiDaS network developed in \cite{Ranftl2021} as the teacher network. For the student architecture, we employed the same architecture from \cite{aleotti2020real}. The student network output was then regressed to match the output of the MiDaS network on our curated dataset. 
% {\color{red}In an effort to further reduce inference times, we ran the NetAdapt pruning tool developed in \cite{cvpr_2021_yang_netadaptv2} on our student network. NetAdapt reduces runtimes by analyzing the device-specific performance of a network and subsequently pruning off sections of the network that perform redundant computations. The use of NetAdapt on our student network resulted in a 2X speedup in inference times on our NVIDIA Jetson Xavier embedded GPU device. } 
Finally, we were able to achieve a 2X inference time speedup by converting our pretrained student network into the Tensor RT framework offered by NVIDIA. The final model runs at an average inference time of 25ms when no other processes are running on the board.

\subsubsection{Classical depth estimation}
As noted before, the outputs of the MiDaS network, and our trained student network, contain estimates of disparity rather than metric depth. Since disparity is equal to metric depth up to shift and scale, obtaining metric depth amounts to obtaining the shift and scale parameters which transform disparity to depth. In MR-iFLY, we employ classical triangulation and re-projection algorithms based on intrinsic properties of the UAS's onboard camera to estimate the shift and scale parameters.

Estimating the shift and scale parameters first requires sparse estimates of metric depth obtained by treating consecutive frames in the UAS's camera stream as a pseudo-stereo system. We use ORB feature matching algorithms~\cite{ORB} implemented in OpenCV to ascertain matching key-point locations in consecutively captured frames and use Lowe's ratio test to remove poor quality matches. The top $n=16$ matches are used in triangulation and re-projection error minimization. The triangulation is done in homogeneous coordinates; given a feature location $p = [x_l, y_l, 1]$ in the first image and its matching location in the second image $p' = [x_r, y_r, 1]$, we wish to ascertain the scene point observed through these pixels, $P = [x, y, z]$. We construct the matrix:
\begin{align}
    A = \begin{bmatrix}
      x_lM_3 - M_1\\
      y_lM_3 - M_2\\
      x_rM'_3 - M'_1\\
      y_rM'_3 - M'_2
\end{bmatrix}
\end{align}
where $M,M'$ are the camera matrices with both intrinsic and extrinsic parameters for the ``left" and ``right" images, respectively. The expressions for the above matrix $A$ are generated by enforcing epipolar geometric constraints on the scene point $P$~\cite{cs231A}. The triangulation for the scene point $P$ is done by solving $AP = 0$ for $P$ using singular value decomposition (SVD). The above method provides an initial solution for the depth of the scene point, which is further refined by minimizing the re-projection error of the computed $P$. This is done by solving the following minimization problem:
\begin{align}
    \minopt_{\hat{P}}~~\norm{M\hat{P} - p}^2 + \norm{M'\hat{P} - p'}^2
\end{align}
In our implementation, the solution for $\hat{P}$ is obtained using the Trust Region Reflective algorithm for non-linear least squares. We employ the soft-L1 loss function:
\begin{align}
    loss(x) = 2(\sqrt{1+x} - 1)
\end{align}
for its robustness to outlier values. The final depth value obtained from the re-projection error minimization problem is subsequently used in the scale parameter estimation. Each of the depth values for the $n$ selected scene points is obtained using the above methods. Predicted disparity values are then scaled separately for each estimated depth and the average over scaled disparities is taken as the metric depth map.

To ascertain the minimum depth in the scene, we use the pixel location of the minimum predicted disparity and find the corresponding pixel location of the minimum disparity scene point in the "right" image using a sliding normalized 2-norm distance filter, selecting the pixel location in the "right" image that yields the minimum filter distance. These pixel locations from the "left" and "right" images are then used in the triangulation and re-projection error minimization procedures. Once the estimated minimum depth is obtained, we add it to the scaled disparity map to obtain the final estimated metric depth. In practice, the minimum and maximum depths obtained via the described estimation procedures can still be quite noisy due to camera sensor and pose sensor errors. Accordingly, we maintain a windowed average of the most recent 6 estimations for these values rather than using the direct estimate at each frame.

% \begin{figure*}[!t]
% \centering
% \subfigure[]{\includegraphics[width=1.1 \columnwidth]{drone_imgs.pdf}
% % \caption{NXP (left) and Crazyflie (right) UAS platforms used for demonstrating the proposed autonomy framework.}
%     \label{fig:drones}}
% \hspace{.2 cm}
% \subfigure[]{\includegraphics[width=.99 \columnwidth]{ex_disp.pdf}
%     % \caption{Examples of disparity estimation (bottom) from RGB (top) and qualitative evaluation features.}
%     \label{fig:ex_disp}}
% \end{figure*}

\begin{figure*}
\begin{minipage}[h]{0.35 \linewidth}
%\vspace{1.3 cm}
\centering
\includegraphics[width=1.0 \columnwidth]{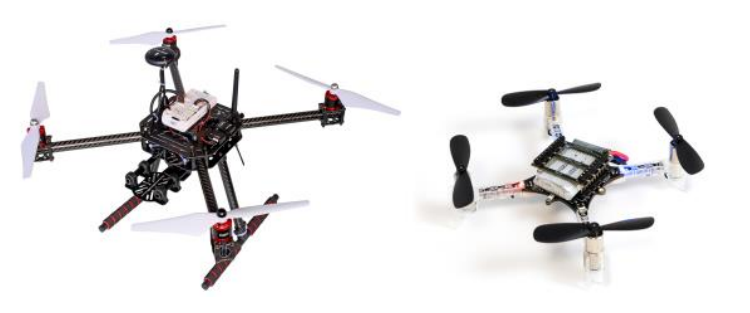}
    \caption{NXP (left) and Crazyflie (right) UAS platforms used for demonstrating the proposed autonomy framework.}
    \label{fig:drones}
\end{minipage}
\hspace{.2 cm}
\begin{minipage}[h]{0.62 \linewidth}
\centering
%\captionof{figure}{Z2 SDR}
\includegraphics[width=.99 \columnwidth]{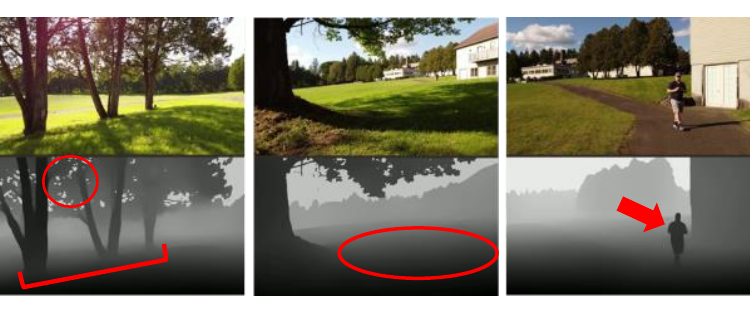}
    \caption{Examples of disparity estimation (bottom) from RGB (top) and qualitative evaluation features.}
    \label{fig:ex_disp}
\end{minipage}
\squeezeu
\end{figure*}

\iffalse
\begin{figure*}[htb]
    \centering
    \includegraphics[width=1.4 \columnwidth]{ex_disp.pdf}
    \caption{Examples of disparity estimation (bottom) from RGB (top) and qualitative evaluation features: arrow shows object continuity, bracket shows relative consistency, ellipses show robustness to interference.}
    \label{fig:ex_disp}
    \squeezeupp
\end{figure*}
\fi

\subsection{Occupancy Grid}
Once the metric depth map has been estimated, an occupancy grid is generated for use in the planning module. The depth map is converted into a point cloud by projecting the 1-dimensional depth values into 3 dimensions using the camera's intrinsic parameters. After the point cloud is constructed, points within a predetermined local vicinity of the agent are added to the occupancy grid by setting a binary flag to denote the presence of an object. The occupancy grid is the collection of these flags and is represented by a binary array. Grid resolution and radius are pre-set algorithmic parameters. 

\subsection{Planning Module}
The occupancy grid is used in the planning module to generate the UAS actions. In MR-iFLY, we use the Fast-Marching Method (FMM)~\cite{sethian} planning algorithm to obtain action sequences. The FMM planner is based on physical models of wavefront propagation and is able to solve high-dimensional planning problems efficiently. To generate a plan, we set a local goal to serve as a destination point. This local goal may be provided by a separate autonomy component, a user or operator, or it may be set statically prior to execution of the UAS autonomy program. If the local goal is outside the radius of the occupancy grid we project the goal point into the occupancy grid. The FMM planner is then used to compute the shortest distance from each point in the occupancy grid to the goal location. The actions along the shortest distance path from the UAS to the goal are recorded.

In our implementation, we commit three actions along the shortest path in each planning cycle. After action execution, a new depth map is obtained and the planning module is run again. Lastly, we pad the occupancy grid with an extra binary flag around each of the obstacle points to restrict the planning of a path that maneuvers the UAS too closely to an obstacle.

\section{Experimental Evaluation and Demonstration}
In this section, we describe the physical implementation and evaluation of MR-iFLY. Due to the nature of the tasks and problems that each subcomponent is employed for, some of the evaluation criteria are both qualitative and quantitative. In either case, we substantiate the validity of the evaluation criteria and subsequent analysis of results.
% In this section we begin by describing the hardware implementation of the proposed framework. We some of the technical specifications of the hardware components and provide images displaying the lab demonstration setup. 

% After detailing the physical implementation, we discuss the various ways we evaluate the overall proposed framework and its constituent subcomponents. Due to the nature of the tasks and problems that each subcompenent is employed for, some of the evaluation criteria are both qualitative and quantitative. In either case, we substantiate the validity of the evaluation criteria and subsequent analysis of results.

\subsection{Hardware Implementation}
To demonstrate the real-world applicability of our proposed framework, we implement and demonstrate the autonomy solution on two separate hardware platforms, depicted in Figure~\ref{fig:drones}.

\iffalse
\begin{figure}
    \centering
    \includegraphics{drone_imgs.pdf}
    \caption{NXP (left) and Crazyflie (right) UAS platforms used for demonstrating the proposed autonomy framework.}
    \label{fig:drones}
\end{figure}
\fi

The first platform, the Bitcraze Crazyflie 2.1 drone, features a lightweight frame measuring 9 cm by 9 cm, a 27-gram unloaded weight, and the ability to attach expansion modules called decks. We equip the Crazyflie with the Flow-v2 deck which provides a relative positioning system and a first-person view (FPV) camera. A 3D printed mount was designed to hold the FPV camera to the drone. Connection to the Crazyflie is achieved via the Crazyradio module. Bitcraze provides a python API to connect to the Crazyflie and perform both low- and high-level controls.

The second platform, the NXP KIT-HGDRONEK66 quadrotor kit, includes a carbon fiber frame, the Flight Management Unit (FMU), RC receiver, RC transmitter, power module, power distribution board, Electronic Speed Controllers (ESCs), motors, GPS, FMU debug adapter, and propellers. Additionally, the frame contains rails for attaching sensors in a standard mount format. A telemetry radio is added to the platform and the NXP is equipped with the ArduCam High Quality Camera. This sensor is a 12.3 MP camera that supports frame rates of 1920 x 1080 at 60 fps and 4032 x 3040 at 30 fps. The selected version of the camera is the “mini” version that uses has a smaller camera board and lens suitable for use on a UAS.

% \begin{figure}
%     \centering
%     \includegraphics{mounts.pdf}
%     \caption{3D-printed mounts for the monocular camera (left), monocular camera and embedded board (middle), stereo camera and embedded board (right).}
%     \label{fig:mounts}
% \end{figure}

To run our algorithms, we utilize both the NVIDIA Jetson Nano and Jetson Xavier embedded GPU development boards. For the Crazyflie platform, algorithmics and actions are computed on the board and sent over a wireless link to be executed on the UAS. For the NXP UAS, we attached the embedded GPU boards to the UAS' frame via custom mounts.
% Images depicting these mounts are provided in Figure \ref{fig:mounts}. 
In Table \ref{spec_table}, we provide technical specifications of the embedded devices.

\begin{table}[!htb]
\centering
\caption{Computational performance specification comparison between NVIDIA Jetson Nano and NVIDIA Jetson Xavier}
% \begin{tabular}{ |c|c|c|c|c| } 
%  \hline
%  Spec. & CPU Cores (\#) & RAM (GB) & AI Performance & GPU Cores (\#) \\ \hline
%  Nano & 4 & 4 & 472 (GigaFlops/s) & 128 \\ \hline
%  Xavier & 6 & 8 & 21 (TensorOps/s) & 384, 48 Tensor Cores
%  \hline
%  \label{spec_table}
% \end{tabular}

\begin{tabular}{ ccc } 
 \toprule
 Spec. & Nano & Xavier \\
 \midrule
 CPU Cores (\#) & 4 & 6 \\
%  \hline
 RAM (GB) & 4 & 8 \\
%  \hline
 AI Performance & 472 (GigaFlops/s) & 21 (TensorOps/s)\\
%  \hline
 GPU Cores (\#) & 128 & 384, 48 Tensor Cores \\
 \bottomrule
\end{tabular}
\label{spec_table}
\end{table}
\subsection{Disparity Estimation}
The task of disparity estimation has straightforward evaluation criteria in the case when ground truth disparity maps are available. A suitable regression metric, such as mean squared error or mean absolute error, may be used to characterize how well the estimation model fits the disparity data. While this approach to evaluation is theoretically sound, it is often the case that ground truth disparity is not available for the deployment platform, sensors, and environment in real-world implementation scenarios. Accordingly, there exist two options for evaluating the efficacy of the disparity estimation model for a specific real-world task: collect ground truth disparity maps and evaluate the model in the usual way, or manually inspect the output of the model and qualitatively evaluate the model's efficacy. The former may often be infeasible if the appropriate sensors are not available or deployable on the target deployment platform. For these reasons, we now discuss how we qualitatively evaluated disparity estimation models.

Perhaps the most important qualitative evaluation characteristics of a disparity estimation model are object continuity, relative consistency, and robustness to interference. For object continuity, we inspect images to ensure that the disparity assigned to pixels that represent the same object do not jump wildly over many values, and further, that jumps do occur at object boundaries. This is displayed in the rightmost image of Figure~\ref{fig:ex_disp}. The arrow indicates the location of the human in the disparity output and it can be seen that the disparity values assigned to the pixels associated with the human are continuous. Furthermore, the human's boundaries are seen clearly in the disparity output. For relative consistency, we mean that the relative ordering of disparity values assigned to different objects in the scene obeys the depth ordering that we would expect from manual observation. This is clearly observed in the leftmost image in Figure~\ref{fig:ex_disp} with the bracket showing how disparity values decrease for each of the further trees in the scene. Lastly, illumination interference can greatly affect the veracity of the disparity estimation module. For this reason, we desire a model which is robust to such interference. In Figure~\ref{fig:ex_disp}, it can be observed that the disparity model's output is robust to illumination interference in the areas enclosed by the ellipses. The model's output is robust to both the bright light from the sun and the shadow from the tree.

\subsection{Runtimes}
While disparity estimation accuracy is important, it is perhaps equally as important that the model's inference does not incur prohibitive latency. In this section, we report the inference times of the disparity estimation model at various points throughout the model reduction and optimization process. These inference times are recorded in Table~\ref{inf_table}.

\begin{table}[!htb]
\centering
%\caption{Inference times for various models on the NVIDIA Jetson Xavier.}
\caption{Inference times on the NVIDIA Jetson Xavier.}
\begin{tabular}{ ccc } 
 \toprule
 Large MiDaS Py. & Mobile PyDnet Py. (\#) & Mobile PyDnet TRT\\ \midrule
 386ms & 55ms & 25ms \\
 \bottomrule
\end{tabular}
\label{inf_table}
\end{table}

The large MiDaS Py. model is the open-source model proposed in \cite{Ranftl2020} without any reduction or optimization. The Mobile PyDnet Py. employs the model architecture proposed in \cite{aleotti2020real} trained on our own procured data set. This model is the knowledge distillation of the large MiDaS model without any optimizations, running in Python. The Mobile PyDnet TRT model is the final model employed in our framework. It is generated by using the TensorRT hardware-specific optimization framework to optimize the Mobile PyDnet network. 
% The inference for this model is performed using the C++ API. 
As is observed in Table~\ref{inf_table}, the model reduction and optimization techniques employed in our work lead to 15X speedups in model inference, allowing for real-time operation of MR-iFLY.

\subsection{Flight Test}
\begin{figure}[!htb]
    \centering
    \includegraphics[width=.88 \columnwidth]{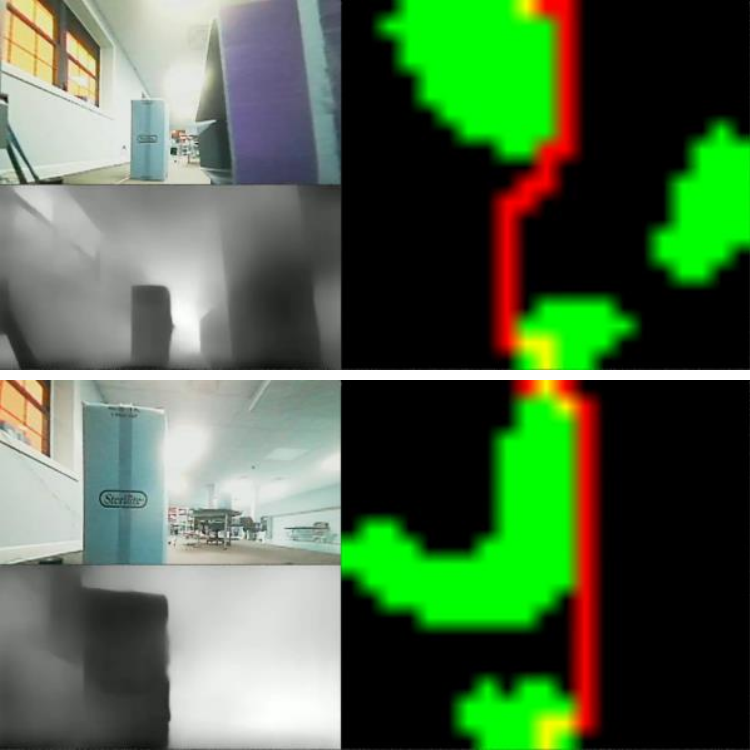}
    \caption{Still frames from the flight test demonstration video. Each frame contains: RGB observation (top left of frame), depth estimation (bottom left of frame), occupancy map (right side of frame, green) and the motion planner's output (red).}
    \label{fig:demo_clips}
\end{figure}
To test the efficacy of the overall solution, we have run demonstration flight tests on basic navigation and collision avoidance tasks. The tasks consist of flying to a goal location behind either a single stack or multiple stacks of boxes. A video of the successful demonstration involving two stacks of boxes provides insight into the inner workings of our framework and is provided at the link in \cite{video}. Figure~\ref{fig:demo_clips} depicts still frames from videos of demonstrations involving two stacks of boxes.  In each frame, there exist three panels displaying the RGB observation (top left), depth estimation (bottom left), and the constructed occupancy map (right side). The 2D occupancy maps in the video represent UAS-level slices from the constructed 3D occupancy grid. Accordingly, they may be readily interpreted as a birds-eye view of the UAS' representation of the environment where the UAS is located in the center of the bottom of the grid. Additionally, the panels on the right side of the frames include the computed paths through the environment towards the goal beyond the boxes. These paths are denoted by the red pixels in the map portion of the frame and provide insight into how the autonomy framework makes decisions about avoiding collisions. These successful flight tests substantiate our lightweight framework as a viable solution to UAS autonomy.

% \begin{figure}
%     \centering
%     \includegraphics{vid_framesv2.pdf}
%     \caption{Still frames from the flight test demonstration video. Each frame has three panels: RGB observation (top left), depth estimation (bottom left), and occupancy map (right).}
%     \label{fig:demo_clipsv2}
% \end{figure}

\section{Related Work}
Due to the multi-faceted nature of the presented work, there exists a multitude of related literature. In this section, we focus on reviewing literature from two divisions of the field. First, we discuss algorithmically relevant works, often coming from the visual simultaneous localization and mapping (VSLAM) community. Secondly, we discuss works focusing on actual hardware implementations of UAS autonomy.

% \subsection{Visual SLAM}
Simultaneous localization and mapping (SLAM) is the task of estimating an autonomous robot's position in the world whilst constructing a map representation of the world in which it is operating. While MR-iFLY is not strictly a SLAM solution - our goal is not to construct a globally consistent map of the environment - many of the autonomy components overlap with the components employed in SLAM solutions. This is especially true in Visual-SLAM solutions, where the primary sensors employed to solve the SLAM task are RGB cameras. For these reasons, we cover V-SLAM works with similar components in this subsection.

One of the most popular V-SLAM algorithms is that of ORB-SLAM \cite{orbslam}, which employs a feature-based key-point selection algorithm for use in constructing a triangulation system to discern pose and perform environment mapping. The major difference between MR-iFLY and the components of ORB-SLAM is that we aim to produce dense depth maps with the aid of a machine learning-based depth estimation module (which utilizes a very sparse triangulation method) while ORB-SLAM performs sparse depth estimation without the use of any machine learning. In recent work, LIFT-SLAM \cite{liftslam}, machine learning based feature detection is used to compute key-point locations that are passed into the V-SLAM pipeline. The authors of \cite{liftslam} demonstrate improved results on test data sets and provide a solution that is a hybrid between machine learning and traditional autonomy components. Therefore, the design concepts of MR-iFLY and \cite{liftslam} are similar in nature. Active neural V-SLAM \cite{chaplot2020learning} employs a similar design. The authors propose a machine learning and traditional autonomy hybrid architecture to perform active SLAM - controlling a UAS with the specific intent of constructing a map of the environment. Their work employs a similar depth estimation and planning module, however; they only consider a 2-dimensional internal environment representation rather than the 3-dimensional representation in our work. Lastly, each of these works differs from ours in that they are only evaluated on a static data set or in simulation, rather than actually implemented on hardware which is a daunting task.

% \subsection{Implementations}
Works with UAS hardware implementations are also abundant. In \cite{kaufmann2018}, the authors introduce a UAS autonomy framework for drone racing that employs both machine learning based and traditional analytic autonomy algorithms. The authors train a vision system to recognize gates and output the location of the desired goal based on gate location. In their follow-up work \cite{loquercio2019}, the authors improve upon their solution by training the vision system completely in simulation to avoid the necessity of collecting large real-world data sets. They show the success of their zero-shot sim-to-real transfer technique in a real-world drone racing setup on an in-house quadcopter. While these works are similar to ours in design philosophy - both works employ a combination of machine learning and analytic autonomy - there are fundamental differences that separate our framework. First, in our work, we focus on robust collision avoidance rather than on optimizing trajectories through gates. Accordingly, the real-world experiments conducted by the authors of \cite{loquercio2019} are set up in an open environment, free of obstacles. Secondly, our work builds an internal representation of the environment via occupancy grid construction, which may be extensible to autonomy tasks beyond collision avoidance, such as exploration, semantic question answering, etc. In \cite{loquercio2019}, only trajectories are computed without constructing a world representation.

In \cite{pedro21}, the authors introduce a machine learning based collision avoidance system for small UAS. The proposed framework utilizes optical flow input to track dynamic obstacles through the environment and subsequently avoid collisions. The authors test their methods on a commercial UAS with commercial vision sensors and demonstrate that the UAS successfully avoids a ball thrown toward the hovering UAS. These experiments differ from ours in that they explicitly consider non-stationary obstacles with a stationary UAS. In our work, we focus on achieving UAS movement whilst avoiding collisions with stationary obstacles.

\section{Conclusion and Future Work}
In this work, we have presented our novel, modular solution to UAS autonomy, entitled MR-iFLY, which leverages techniques from both machine-learning and traditional autonomy algorithms. We have described, in detail, the workings of the algorithmic components of MR-iFLY and have elucidated how each component interacts with one another. We have described the various evaluation criteria we have used to measure performance and shown how our optimized machine vision components are robust to environmental factors and provide up to 15X speedup over the original model. Further, we have provided a demonstration video highlighting a successful flight test on a UAS with an embedded GPU board. These empirical results, along with our design philosophy, meet and mitigate the demands of vision-based autonomous UAS problems associated with various industrial applications. Accordingly, we believe that MR-iFLY may be leveraged by 6G communications swarms to reduce communication overhead between nodes by providing standalone collision avoidance and navigation capabilities.

In future work, we plan to extend MR-iFLY in various ways. First, we expect to add a further machine learning component that computes the directive local goals that the present framework will use in planning and navigation. This component is expected to be learned via a reinforcement learning scheme, in which a policy would be learned to output a sequence of waypoints, which together, achieve some overarching mission goal. In addition, we wish to investigate the use of probabilistic planners in the motion planning module. In our current implementation, the FMM planner assumes that obstacles and objects in the planning grid are static, and thus the computed plan may only be valid for a short time over which the static obstacle assumption is reasonable. 
% Also, for missions in which the target point is mobile, a non-deterministic planner would be advantageous so that different possible goal locations may be considered. With regard to the depth estimation module, future work may consist of trying to phase out the sparse keypoint detection part of the algorithm. An envisioned path forward for this is as follows. Using a stereo system in conjunction with an RGB camera, collect a data set of stereo-RGB pairs with which the knowledge distilled disparity network can be fine-tuned Once fine-tunes with the collected stereo data, the networks disparity output should correspond to the disparity associated with the particular hardware employed to collect the data. The disparity output at operation time can then be scaled to metric depth using the usual baseline scaling technique associated with the data collection stereo hardware. 
Lastly, we wish to employ MR-iFLY in a swarm mission, and test to see if each of the UAS in the swarm can avoid each other without the need for communication, thereby further substantiating MR-iFLY as a potential solution for autonomy in 6G communications swarms.

\bibliographystyle{ieeetr}
\bibliography{mrlab}

\end{document}